\documentclass{article} % For LaTeX2e
\usepackage[accepted]{icml2023}
\usepackage{times}
\usepackage{comment}

\usepackage{amsmath,amsfonts,bm,amssymb}

\def\eqref#1{equation~\ref{#1}}
\def\1{\bm{1}}

\def\va{{\bm{a}}}

\def\vh{{\bm{h}}}

\def\vm{{\bm{m}}}

\def\vv{{\bm{v}}}

\def\vx{{\bm{x}}}

\def\mV{{\bm{V}}}

\def\mX{{\bm{X}}}

\DeclareMathAlphabet{\mathsfit}{\encodingdefault}{\sfdefault}{m}{sl}
\SetMathAlphabet{\mathsfit}{bold}{\encodingdefault}{\sfdefault}{bx}{n}

\renewcommand{\Re}{\mathbb{R}}

\usepackage{adjustbox}

\usepackage{hyperref}
\usepackage{url}
\usepackage{booktabs}
\usepackage{multicol}
\usepackage{subcaption}

\icmltitlerunning{Using Multiple Vector Channels Improves E(n)-Equivariant Graph Neural Networks}

\begin{document}

\twocolumn[
\icmltitle{Using Multiple Vector Channels Improves E(n)-Equivariant\\ Graph Neural Networks}

\icmlsetsymbol{equal}{*}

\begin{icmlauthorlist}
\icmlauthor{Daniel Levy}{equal,mcgill,mila}
\icmlauthor{Sékou-Oumar Kaba}{equal,mcgill,mila}
\icmlauthor{Carmelo Gonzales}{intel}
\icmlauthor{Santiago Miret}{intel}
\icmlauthor{Siamak Ravanbakhsh}{mcgill,mila}
\end{icmlauthorlist}
\icmlaffiliation{mcgill}{School of Computer Science, McGill University, Montreal, Canada}
\icmlaffiliation{mila}{Mila, Quebec AI Institute}
\icmlaffiliation{intel}{Intel Labs}

\icmlcorrespondingauthor{Daniel Levy}{daniel.levy@mila.quebec}
\icmlcorrespondingauthor{Sékou-Oumar Kaba}{kabaseko@mila.quebec}

\icmlkeywords{Machine Learning, ICML}

\vskip 0.3in
]

\printAffiliationsAndNotice{\icmlEqualContribution} % otherwise use the standard text.

\begin{abstract}
We present a natural extension to E($n$)-equivariant graph neural networks that uses multiple equivariant vectors per node.
We formulate the extension and show that it
improves performance across different physical systems benchmark tasks, with minimal differences in runtime or number of parameters. The proposed multi-channel EGNN outperforms the standard single-channel EGNN on N-body charged particle dynamics, molecular property predictions, and predicting the trajectories of solar system bodies.
Given the additional benefits and minimal additional cost of multi-channel EGNN, we suggest that this extension may be of practical use to researchers working in machine learning for the physical sciences.
\end{abstract}

\section{Introduction}

Designing neural network architectures that correctly account for the symmetry of physical laws is an important requirement for applications of artificial intelligence in science.
In particular, for dynamical systems in physics, the relevant properties transform invariantly or equivariantly under Euclidean transformations. This is also the case when modelling particles or atomistic systems, for which machine learning simulators are already widely used.

There now exist a wide variety of equivariant neural network architectures leveraging a diverse set of mathematical formulations, among which we highlight two important classes.
First, some architectures apply spherical harmonics mappings to incorporate directional information in an equivariant way \citep{thomas2018tensor, fuchs2020se3transformers}.
These architectures have the advantage of being highly expressive but are also computationally expensive and challenging to implement.
Second, we highlight models that fall under the equivariant multilayer perceptron (E-MLP) paradigm \citep{finzi2021practical}.
The idea behind this paradigm is to simply generalize standard multilayer perceptrons by composing equivariant linear layers with appropriate non-linear functions.
These architectures are much simpler to work with and more computationally efficient than those using spherical harmonics, but in principle, require lifting input quantities to high-order tensors to achieve high expressivity \citep{finkelshtein2022simple}.
Prior works, however, show that one can achieve satisfactory modelling performance without requiring higher-order representations. 
One such example is the Vector Neurons \citep{deng2021vector} model, which can be seen as an equivariant multilayer perceptron with order-1 vector features.
This architecture also leverages the fact that the number of vector channels (neurons) in each layer can be arbitrary to increase expressivity.

The E($n$)-equivariant Graph Neural Network (EGNN) model by \citet{satorras2021n} is an example of a model that does not clearly fit into one of the categories above.
Nevertheless, EGNN has become widely applied mainly due to efficiency and simple model design. 
EGNN uses the message-passing framework, which captures the inductive bias that sparse interactions between entities should lead to better generalization.
EGNN also has the advantage of separating equivariant features into a separate channel that only follows equivariant operations.
The work of \cite{brandstetter2022geometric} extends EGNN by using ideas inspired by spherical-harmonics-type architectures.
Their Steerable E($n$)-Equivariant Graph Neural Network (SEGNN) achieves better performance across some benchmarks but suffers from similar conceptual shortcomings in addition to increased computational complexity.

\begin{figure*}[t]
\centering
    \begin{subfigure}[t]{0.18\textwidth}
        \centering
    \includegraphics[height=1.5\linewidth]{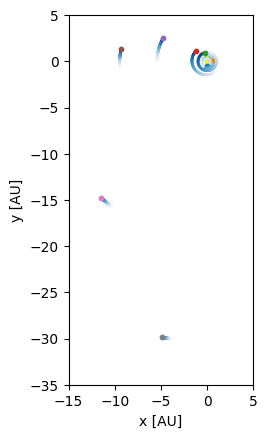}
        \caption{Full solar system}
    \end{subfigure}%
    ~
    \begin{subfigure}[t]{0.36\textwidth}
        \centering
    \includegraphics[height=0.75\linewidth]{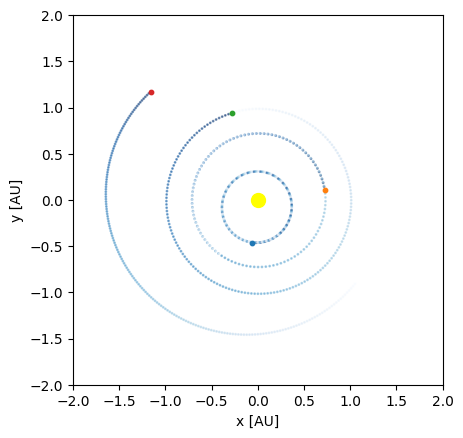}
        \caption{Close-up of the inner solar system}
    \end{subfigure}%
    ~
    \begin{subfigure}[t]{0.36\textwidth}
        \centering
     \includegraphics[height=0.75\linewidth]{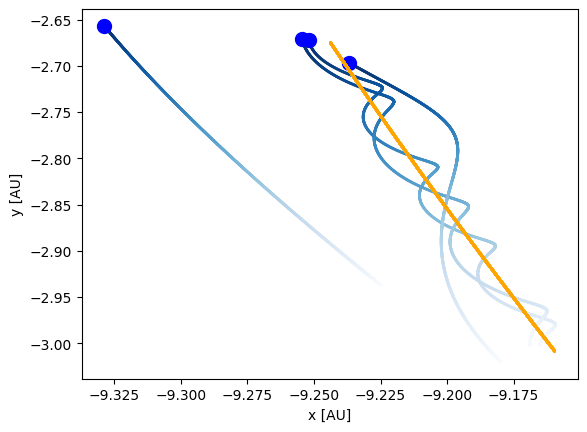}
        \caption{Close-up on 4 moons of Saturn}
    \end{subfigure}%
\caption{Visualization of the Solar Systems dynamics dataset. Solid circles represent initial positions of celestial bodies and traces their training trajectories.}
\label{fig:solar}
\end{figure*}

In this paper, we explore the direction of generalizing EGNN by drawing from E-MLP-type architectures.
EGNN only updates a single vector for each node in the graph over each layer.
A natural way to increase the expressivity of this model is to make the number of vector channels arbitrary.
In our experiments, we show that this change alone leads to an important increase in performance for some physical modelling tasks.
This multi-channel extension also retains the simplicity and computational efficiency of the original architecture and makes intuitive physical sense:
the network may use the different channels to store additional physical quantities relevant to the prediction task.

We note that GMN \citep{huang2022equivariant} proposes to use multiple channels as part of a generalized EGNN-like model, as does GVP-GNN \citep{jing2021learning}.
However, since this is one contribution amongst several others in GMN, and GVP-GNN the advantage of using multiple channels is not clear. Here we show that we can obtain significant benefits only with the additional channels.
In this short paper, we highlight that simply adding multiple channels to EGNN can lead to a significant performance increase compared to much more expensive methods such as SEGNN.
We believe this result should be of use to practitioners looking to preserve the advantages of EGNN.

\section{Method}

Following the formulation of EGNN, we assume that the model operates on graphs embedded in $n$-dimensional Euclidean space (typically $n=3$). To each node is associated coordinates $\vx_{i} \in \Re^n$ and node features $\vh_{i} \in \Re^{d_h}$. Edge features $\va_{i,j} \in \Re^{d_e}$ can also be considered. The original EGNN layer is defined by the following equations:
\begin{align}
    \vx_{ij} &= \vx_{i} - \vx_{j}
\\
    \vm_{ij} &= \boldsymbol\phi_e\left(\vh_{i}, \vh_{j}, ||\vx_{ij}||^2, \va_{ij}\right)
\\
    \vx_{i}^{t+1} &= \vx_{i}^{t} + C \sum_{j\in\mathcal{N}(i)} \vx_{ij}^t\phi_x(\vm_{ij})
\\
    \vh_{i}^{t+1} &= \boldsymbol\phi_h(\vh_{i}^{t}, \sum_{j\in\mathcal{N}(i)}\vm_{ij}) \label{eq:hidden_update}
\end{align}
where
$\boldsymbol\phi_e: \Re^{d_h + d_h + 1 + d_e} \rightarrow \Re^{d}$,
$\phi_x: \Re^{d} \rightarrow \Re$
and
$\boldsymbol\phi_h: \Re^{d_h + d} \rightarrow \Re^{d_h}$ are multilayer perceptrons (MLPs) and $\mathcal{N}(i)$ is the neighborhood of node $i$.

We define the Multi-Channel E($n$)-Equivariant Graph Neural Network (MC-EGNN) by 
replacing $\vx_i$ with the matrix 
$\mX_{i} \in \Re^{3 \times m}$, where $m$ is the number of vector channels.
Using this, we modify the above equations as follows:
\begin{align}
\mX_{ij} &= \mX_{i} - \mX_{j}
\\
%\vr(\mX) &=  [||\mX_{:,1}||^2, \dots ||\mX_{:,m}||^2]^\top \label{eq:dist_calc}
%\\
    \vm_{ij} &= \boldsymbol\phi_e\left(\vh_{i}, \vh_{j}, ||\mX_{ij}||_{\text{c}}^{2}, \va_{ij}\right)
\\
    \mX_{i}^{t+1} &= \mX_{i}^t + C \sum_{j\in\mathcal{N}(i)} \mX_{ij}^t\Phi_x(\vm_{ij}) \label{eq:channel_pos_update}
\end{align}
for MLPs
$\boldsymbol\phi_e: \Re^{d_h + d_h + m + d_e} \rightarrow \Re^{d}$
and
$\boldsymbol\Phi_x: \Re^{d} \rightarrow \Re^{m \times m'}$, where $m'$ is the output channel dimension. $||\mX_{ij}||_{\text{c}}$ denotes the channel-wise Euclidean norm. 
Equation \ref{eq:hidden_update} stays the same.
We set $m=1$ for the first and last layer to use a single vector for each node's inputs and outputs.
The modification to EGNN does not affect the equivariance properties of the architecture since the Euclidean and permutation groups do not act over the channels dimension.

\section{Experiments}

\subsection{Solar System Predictions}
\subsubsection{Dataset}
To investigate the necessity of different number of vector channels, we perform experiments on prediction of dynamics on an N-body system based on the solar system.
We look at a dataset of 30 years of real observations of the 31 highest-mass bodies in the solar system (including the sun, 8 planets, and 22 moons) curated by \citet{lemos2022rediscovering} and sourced originally from NASA Horizons \footnote{https://ssd.jpl.nasa.gov/horizons/}.
We set up the task of simultaneously predicting the positions of each of the bodies 60 Earth-days into the future, given each of their positions, velocities, and log-masses.

We constructed a train/validation/test split using three subsequent years of data, with one year of data for each partition. We provide a visualization of sample training set trajectories at Figure \ref{fig:solar}.
We trained our models on a fully connected graph of the solar system bodies using a mean squared error loss normalized by the true distance between the initial and final positions of each bodies. This is done to account for the broad ranges in velocities in the system.

\begin{figure}[h!]
\centering
    \includegraphics[width=0.75\linewidth]{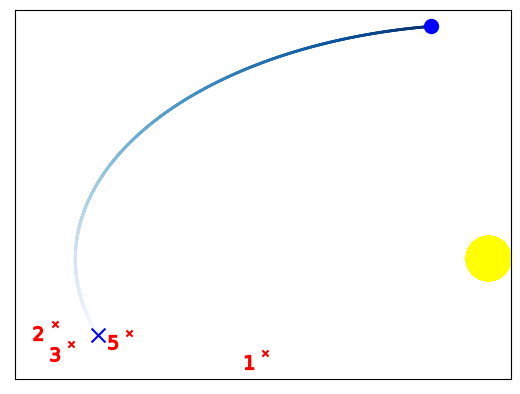}\par
\caption{Predictions of the position of Venus for different numbers of vectors channels. The cross indicates the ground truth position.}
\label{fig:prediction}
\end{figure}

\subsubsection{Results}
Most of the predictions made in this task involve moons orbiting planets while also orbiting the sun, and so we hypothesized that 3 vectors would be needed to approximate their dynamics efficiently: to keep track of the coordinates, the angular momentum around the sun, and the angular momentum around the planet. Note that velocity is already considered since we use a variant of the multi-channel EGNN described in Appendix \ref{sec:vel}.

We first conducted a hyperparameter search using a one vector-channel EGNN to maximize its performance on the validation set, and then used those hyperparameters when testing the EGNN models with 2, 3, and 5 vector channels. 

Our results, shown in Table \ref{tab:solar}, validate our hypothesis.
While using 2 vector channels improves over using 1, it takes 3 vector channels for the model to achieve its highest performance. Note that the difference in performance between using 3 and 5 channels is not statistically significant, it is therefore not crucial to tune this parameter to an exact value. Figure \ref{fig:prediction}, shows clearly that original single channel EGNN model is not able to provide an accurate estimate of the future and is widely off the trajectory.

\begin{table}[h!]
  \centering
  \caption{Performance on the solar system prediction task using differing number of vector channels. Performance is shown averaged across all 31 solar system bodies, and normalized by true distance.}
    \vspace{1ex}
    \footnotesize
  \begin{adjustbox}{width=.3\textwidth}
  \begin{tabular}{ll}

  \toprule

\# of Vectors & Normalized MSE \\
\midrule
1                 & 0.109 $\pm$ 0.051       \\
2                 & 0.082 $\pm$ 0.047       \\
\bf{3}            & \bf{0.024 $\pm$ 0.007}       \\
5                 & 0.030 $\pm$ 0.008       \\
    \bottomrule
  \end{tabular}
  \end{adjustbox}
    \vspace{-1ex}
  \label{tab:solar}
\end{table}

\subsection{Charged Particles System}

We also compare againt other models using a widely used benchmark.
In the Charged Particles N-body experiment \citep{kipf2018neural}, the task is to predict the positions of charged particles several timesteps into the future, given their charges, positions, and velocities.

We use the variant created by \cite{satorras2021n}, which consists of 3,000 training samples, each consisting of a system of 5 particles with their charges and 3d coordinates and velocities given, and we train our network on a loss of the mean squared error of the particle's position after 1000 timesteps.
We also use the velocity version of the multi-channel EGNN in this experiment.
For this experiment, we use the implementation and hyperparameters of the EGNN used by \cite{satorras2021n}.
The only modification we make is to incorporate multiple vector channels.
Specific architecture details and hyperparameters are listed in Appendix \ref{sec:hp_nbody}.

\begin{table}[h]
    \centering
    \caption{
    Test set MSE for the N-body experiment.
    Our results are averaged over 5 random seeds.
}
    \vspace{1ex}
    \footnotesize
    \begin{adjustbox}{width=0.5\textwidth}
    \begin{tabular}{l l}
  \toprule
    Method & MSE  \\
    \midrule
    SE(3)-Tr \citep{fuchs2020se3transformers} & 0.0244 \\
    TFN \citep{thomas2018tensor} & 0.0155 \\
    NMP \citep{gilmer2017} & 0.107 \\
    Radial Field \citep{kohler2019equivariant} & 0.0104 \\
    CN-GNN \citep{kaba2022equivariance} & 0.0043 $\pm$ 0.0001 \\
    FA-GNN \citep{puny2022frame}       & 0.0057 $\pm$ 0.0002  \\
    SEGNN  \citep{brandstetter2022geometric}     & 0.0043 $\pm$ 0.0002  \\
    \midrule
    EGNN \citep{satorras2021n}       & 0.0070 $\pm$ 0.0005  \\
    \textbf{MC-EGNN-2}  & $\bf{0.0041 \pm 0.0006}$  \\
    MC-EGNN-5  & 0.0043 $\pm$ 0.0003  \\
    MC-EGNN-10 & 0.0044 $\pm$ 0.0005 \\
    MC-EGNN-25 & 0.0048 $\pm$ 0.0005  \\
    \bottomrule \\
    \vspace{-1.1cm}
  \end{tabular}
  \end{adjustbox}
    \label{tab:n-body}
\end{table}

The results, shown in Table \ref{tab:n-body}, demonstrate that using just 1 more vector channel (MC-EGNN-2) yields greatly improves performance over the single-vector EGNN and matches the performance of more sophisticated models such as SEGNN \citep{brandstetter2022geometric}.
This is at a negligible added cost: Table \ref{tab:n-body-param} (Appendix \ref{apd:comp}) shows that for a small number of vector channels, the forward time and the model's number of parameters are largely unaffected.

\begin{table*}[h!]
  \centering
  \caption{Mean absolute error for property prediction on QM9.}
    \vspace{1ex}
  \begin{adjustbox}{width=1.\textwidth}
  \begin{tabular}{l c c c c c c c c c c c c c c c c c c}
  \toprule
    Task & $\alpha$ & $\Delta \varepsilon$ & $\varepsilon_{\mathrm{HOMO}}$ & $\varepsilon_{\mathrm{LUMO}}$ & $\mu$ & $C_{\nu}$ & $G$ & $H$ & $R^2$ & $U$ & $U_0$ & ZPVE \\
    Units & bohr$^3$ & meV & meV & meV & D & cal/mol K & meV & meV & bohr$^3$ & meV & meV & meV \\
    \midrule
    NMP~\citep{gilmer2017} & .092 & 69 & 43 & 38 & .030 & .040 & 19 & 17 & .180 & 20 & 20 & 1.500 \\
    SchNet~\citep{schutt2017schnet} * & .235 & 63 & 41 & 34 & .033 & .033 & 14 & 14 & .073 & 19 & 14 & 1.700 \\
    Cormorant~\citep{anderson2019cormorant} & .085 & 61 & 34 & 38 & .038 & .026 & 20 & 21 & .961 & 21 & 22 & 2.027  \\
    L1Net~\citep{miller2020relevance} & .088 & 68 & 46 & 35 & .043 & .031 & 14 & 14 & .354 & 14 & 13 &  1.561\\
    LieConv~\citep{finzi2020generalizing} & .084 & 49 & 30 & {25} & .032 & .038 & 22 & 24 & .800 & 19 & 19 & 2.280 \\
    TFN~\citep{thomas2018tensor} & .223 & 58 & 40 & 38 & .064 & .101 & - & - & - & - & - & - \\
    SE(3)-Tr.~\citep{fuchs2020se3transformers} & .142 & 53 & 35 & 33 & .051 & .054 & - & - & - & - & - & -  \\
    DimeNet++~\citep{klicpera2020fast} * & .0435 & 32.6 & 24.6 & 19.5 & .0297 & .0230 & 7.56 & 6.53 & .331 & 6.28 & 6.32 & 1.21 \\
    SphereNet~\citep{liu2022spherical} * & .0465 & \bf{32.3} & \bf{23.6} & \bf{18.9} & .0269 & \bf{.0215} & 8.0 & 6.40 & .292 & 7.33 & 6.26 & \bf{1.12}\\
    PaiNN~\citep{schutt2021equivariant} * & \bf{.045} & 45.7 & 27.6 & 20.4 & \bf{.012} & .024 & \bf{7.35} & \bf{5.98} & \bf{.066} & \bf{5.83} & \bf{5.85} & 1.28\\
    SEGNN~\citep{brandstetter2022geometric} & .060 & 42  & 24 & 21 & .023 & .031 & 15 & 16 & .66 & 13 & 15 & 1.62 \\
    EGNN~\citep{satorras2021n} & {.071}  & {48} & {29} & {25} & .029 & .031 & 12 & 12 & .106 & 12 & 12 & 1.55  \\
    \midrule
    MC-EGNN-8 & .063 & 42 & 24 & 23 & .016 & .028 & 10 & 11 & .101 & 11 & 10 & 1.45 \\
    \bottomrule
    & \multicolumn{12}{l}{* these methods use different train/val/test partitions.}
  \end{tabular}
  \end{adjustbox}
  \label{tab:qm9}
\end{table*}

\subsection{QM9 – Molecular Property Prediction}

Lastly, we applied the EGNN the task of predicting chemical properties of small molecules.
The QM9 dataset consists of 100,000 training samples of molecules, each described by the atom type and positions of their constituent atoms \citep{ramakrishnan2014quantum}. We use the same training setup as EGNN which facilitates comparison. The hyperparameters are detailed in Appendix \ref{apd:qm9}.

We predict 12 chemical properties using the multi-channel EGNN. In theory, the properties are entirely determined by the atom types and positions.  Unlike in the N-body experiment or the solar system experiment, the predicted properties are coordinate-invariant. The output is therefore obtained by pooling the invariant node embeddings. The vector channels are only used in the intermediate layers, but as we report hereafter, they still contribute to increased performance on all targets compared to the standard EGNN. The results, shown in Table \ref{tab:qm9}, demonstrate that performance is also comparable to SEGNN when using 8 vector channels.

\section{Conclusion}
We show here that adding multiple channels to the EGNN model leads to performance improvements in prediction tasks on physical systems, sometimes matching more complicated architectures. This is achieved without a significant increase in the forward runtime of the model because only a small number of vector channels are needed to obtain improvements.

This generalization could also be useful for tasks where the number of input vectors attached to each node or to predict is arbitrary (for example, when quantities such as angular velocity, spin, or polarization are included). Translationally invariant predictions can also be produced simply by removing the residual connection in the positions update equation.

We plan to investigate further whether there is a particular semantic meaning to the different vectors computed by the multi-channel model that makes it helpful.
One possible downside we noticed with the multi-channel model was that training could be less stable when more vector channels were used.
In practice, we found that gradient clipping could be used to help address this issue, but this was not used in our experiments.
Analyzing the learned vectors could lead to better architecture design and hyperparameter selection.

We further plan to apply this model to larger datasets where we may have many more interactions than in any of the experiments tested in this paper.
We believe that the relative computational efficiency of the method proposed here may allow it to prove useful in applications that were previously unattainable for E($n$) equivariant neural networks.

\bibliography{bib}
\bibliographystyle{icml2023}

\clearpage
\appendix
\section{Velocity Variant} \label{sec:vel}
For problems in which we are provided with node velocities, a similar reformulation of the model to \cite{satorras2021n} is used.
This is done simply by replacing Equation \ref{eq:channel_pos_update} with the following:

\begin{align}
    \mV_{(i)}^{t+1} &= \vv_{(i)}^0 \boldsymbol\phi_v(\vh_{(i)}^t) + C \sum_{j \neq i} \mX_{(ij)}\Phi_x(\vm_{(ij)})
\\
\mX_{(i)}^{t+1} &= \mX_{(i)}^t + \mV_{(i)}^{t+1}
\end{align}

Where $\boldsymbol\phi_v \in \Re^{d} \rightarrow \Re^{m'}$ is an MLP that weighs the impact of the initial velocity vector on the position update.

\section{Additional Results}

\subsection{Comparison of forward time and parameters}
\label{apd:comp}

We show hereafter a comparison of the forward time of the model as well as the number of parameters for different values of number of vectors.

\begin{table}[h]
    \centering
    \caption{
    Comparison of the forward time (for a batch of 100 samples) and number of parameters of EGNN models with different numbers of vector channels used for the N-body experiment. 
}
    % \vspace{1ex}
    \footnotesize
    %\begin{adjustbox}{width=0.45\textwidth}
    \begin{tabular}{l l l}
  \toprule
\# of Vectors & Forward Time (s) & \# of Parameters \\
\midrule
1 (EGNN)            & 0.003118         & 134020               \\
2             & 0.003229         & 135244               \\
5             & 0.003135         & 142012               \\
10            & 0.003405         & 163612               \\
25            & 0.004188         & 305812                 \\     
    \bottomrule \\
    \vspace{-1.1cm}
  \end{tabular}
  %\end{adjustbox}
    \label{tab:n-body-param}
\end{table}

\subsection{Open Catalyst Project – IS2RE}
The Open Catalyst (OC20) Dataset is a large-scale dataset designed for applying machine learning to the task of catalyst discovery, of particular interest to the materials science community \citep{chanussot2021open}.
The dataset contains over 1.2 million simulated molecular relaxations involving different adsorbates and catalytic surfaces.
One challenge associated with the dataset is  the Initial Structure to Relaxed Energy (IS2RE) task, wherein a neural network is trained to predict the relaxation energy of a system described by the initial positions and atom types of a molecule, a catalytic surface, and a bulk crystal substrate below the surface.

We conducted experiments using the Open MatSci ML Toolkit \citep{miret2022open}, which provides an interface into OC20 tasks while abstracting away implementation complexities.
The toolkit also contains an implementation of EGNN, which we modified to include our multi-channel extension.
We used a training set of $\sim$500K training samples and $\sim$25K validation samples and trained on the mean absolute error (MAE) of the relaxed energy for a maximum of 100 epochs, with early stopping.

\begin{table}[h]
    
    \centering
    \caption{
    In-Distribution Validation set MAE for the IS2RE task. EGNN variants was run across 5 random seeds, but due to instabilities in training, seeds where training failed were discarded. MegNet and Gala results are taken from \cite{miret2022open}
    }
    \vspace{1ex}
    % \footnotesize
    \begin{adjustbox}{width=0.45\textwidth}
    \begin{tabular}{l l}
  \toprule
Model      & Validation MAE (eV)   \\
\midrule
MegNet \citep{megnet} & 0.233 \\
Gala \citep{spellings2021geometric} & 0.240 \\
\midrule
EGNN       & 0.239 $\pm$ 0.003 \\
MC-EGNN-2  & 0.262 $\pm$ 0.017                  \\
MC-EGNN-5  & 0.239 $\pm$ 0.006 \\
MC-EGNN-10 & 0.253 $\pm$ 0.019                  \\
    \bottomrule \\
    \vspace{-1.1cm}
  \end{tabular}
  \end{adjustbox}
    \label{tab:is2re}
\end{table}

We report a negative result on the IS2RE task as shown in Table \ref{tab:is2re}. We do not observe any clear benefit with using multiple vectors, with different resulting performances being mostly determined by the random seed used.

\section{Experimental details}
\subsection{N-Body System}
We used the same architecture and hyperparameters as EGNN \cite{satorras2021n} in their N-Body system experiment.
We used 4 EGNN layers. Each layer used 64 channel, 2 layer MLPs for the node, edge, and coordinate update functions, with a Swish function nonlinearity.
We trained on 3000 training samples for 10,000 epochs with a batch size of 100, using the Adam optimizer with a learning rate of $5 \times 10^{-4}$.
\label{sec:hp_nbody}

\subsection{QM9}
\label{apd:qm9}

We used the same architecture and hyperparameters as EGNN \cite{satorras2021n} in their QM9 experiments. Training was performed using the Adam optimizer with learning rate $5\times 10^{-4}$ for all targets except the gap, homo and lumo. We used a cosine learning rate scheduler. Batch size was set at 96 and number of epochs to 1000 with early stopping. We used 8 vector channels and 128 channels for node features and messages. The number of layers was set at 7.

\subsection{Solar System}
We again used the same architecture as EGNN.
We conducted a hyperparameter search using a baseline 1-vector EGNN, looking at layers between 3 to 6, and number of features between $[64, 128, 256, 512]$. We then used a 5-layer EGNN with 128-channels for node features and messages, the Adam optimizer with a learning rate of $1 \times 10^{-4}$ for all experiments with different numbers of vectors. We trained for 1000 epochs with early stopping.

\subsection{Open Catalyst Project}
We used the same EGNN implementation, hyperparameters, and optimization procedure as \cite{miret2022open}.
We used a 3-layer EGNN with 2-layer MLPs for the node, edge, and coordinate update functions, that used 48, 16, and 64 channels respectively.
Node features were first embedded using a 3-layer, 64-channel MLP, and a 3-layer, 128-channel MLP was used to predict relaxed energy using the EGNN's pooled node embeddings.
Training was performed over a maximum of 100 epochs with a with early stopping, with a batch size of 8. The Adam optimizer was used with a learning rate of 1e-5, a gamma of 0.5, and a cosine annealing learning rate scheduler.
\end{document}